\documentclass[conference]{IEEEtran}
\IEEEoverridecommandlockouts
\usepackage{cite}
\usepackage{amsmath,amssymb,amsfonts}
\usepackage{algorithmic}
\usepackage{graphicx}
\usepackage{textcomp}
\usepackage{xcolor}
\usepackage{siunitx}
\usepackage[urlbordercolor=lightgray,hidelinks]{hyperref}
\def\BibTeX{{\rm B\kern-.05em{\sc i\kern-.025em b}\kern-.08em
    T\kern-.1667em\lower.7ex\hbox{E}\kern-.125emX}}
\usepackage[margin = 2cm]{geometry}
\begin{document}

\title{Experimental method for perching\\ flapping-wing aerial robots\\

\thanks{ The research is funded by the European 
Project GRIFFIN ERC Advanced Grant 2017, Action 
788247.}
}

\author{\IEEEauthorblockN{Raphael Zufferey}
\IEEEauthorblockA{
\textit{GRVC Robotics Lab},\\
University of Seville\\
Seville, Spain, and \\
\textit{EPFL}, Switzerland\\
\href{mailto:raph.zufferey@gmail.com}{raph.zufferey@gmail.com} \\
}
\and
\IEEEauthorblockN{Daniel Feliu-Talegón}
\IEEEauthorblockA{\textit{GRVC Robotics Lab}\\
University of Seville\\
Seville, Spain\\
\href{mailto:danielfeliu@us.es}{danielfeliu@us.es}\\
}
\and
\IEEEauthorblockN{Saeed Rafee Nekoo}
\IEEEauthorblockA{\textit{GRVC Robotics Lab}\\
University of Seville\\
Seville, Spain\\
\href{mailto:saerafee@yahoo.com}{saerafee@yahoo.com}\\
}
\and
\IEEEauthorblockN{Jose-Angel Acosta}
\IEEEauthorblockA{\textit{GRVC Robotics Lab}\\
University of Seville\\
Seville, Spain\\
\href{mailto:jaar@us.es}{jaar@us.es}\\
}
\and\IEEEauthorblockN{Anibal Ollero}
\IEEEauthorblockA{\textit{GRVC Robotics Lab}\\
University of Seville\\
Seville, Spain\\
\href{mailto:aollero@us.es}{aollero@us.es}\\
}

}

\maketitle

\begin{abstract}
In this work, we present an experimental setup and guide to enable the perching of large flapping-wing robots. The combination of forward flight, limited payload, and flight oscillations imposes challenging conditions for localized perching. The described method details the different operations that are concurrently performed within the 4 second perching flight. We validate this experiment with a 700 g ornithopter and demonstrate the first autonomous perching flight of a flapping-wing robot on a branch. This work paves the way towards the application of flapping-wing robots for long-range missions, bird observation, manipulation, and outdoor flight.
\end{abstract}


\section{Introduction}

\begin{figure*}[h]
    \centering
    \includegraphics[width=\textwidth]{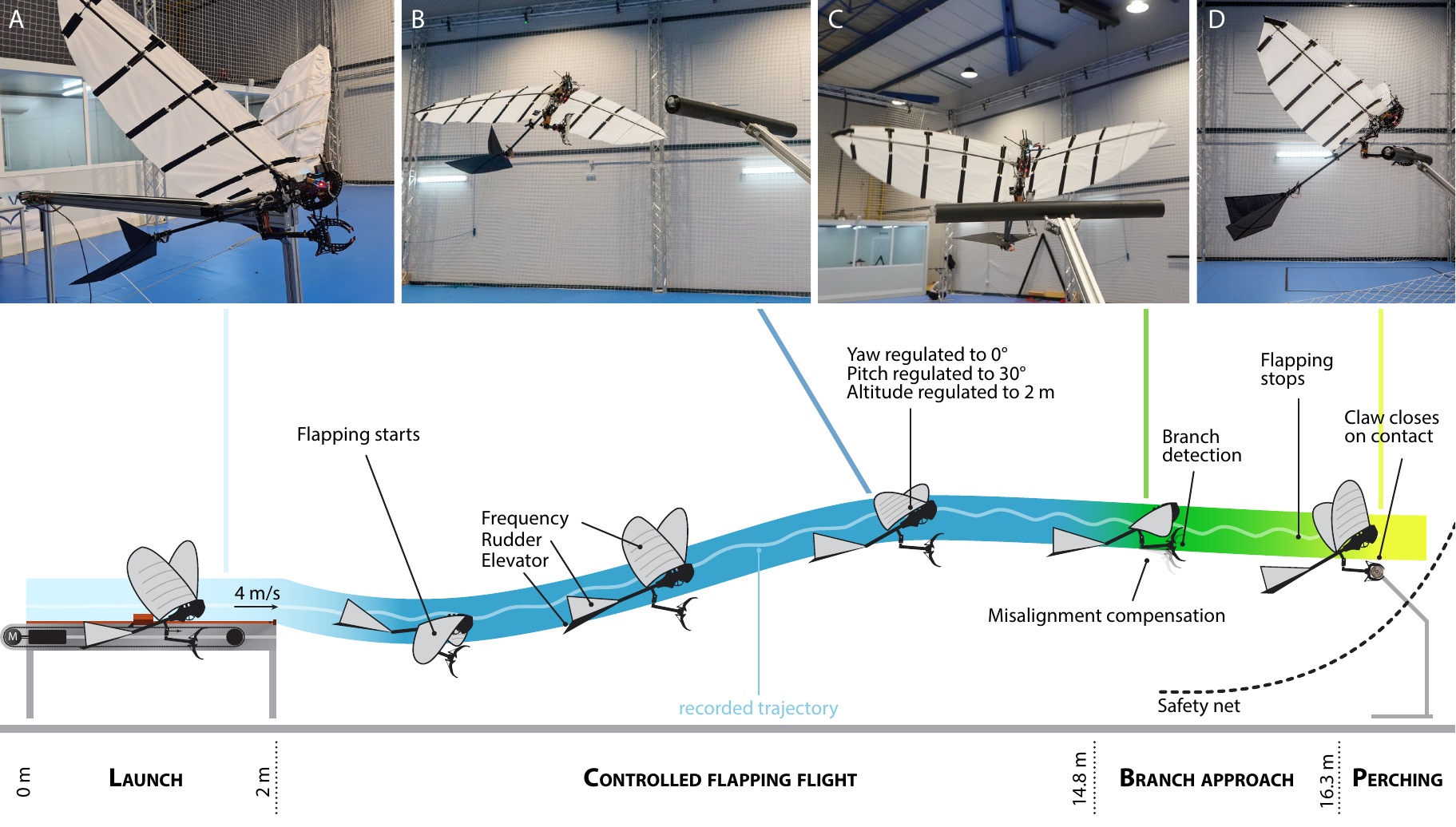}
    \caption{Side view of a typical perching flight experiment. The robotic birds are accelerated by a launching system to 4 m/s. The birds then separate from the launcher and start the controlled flight phase, where the altitude, pitch, yaw and speed is regulated by an autopilot through elevator, rudder and flapping frequency actions. A perching robot is shown in the following positions: \textbf{A.} At the end of the launcher \textbf{B.} In controlled flapping flight \textbf{C.} Shortly before contact, with the misalignment compensation system active and \textbf{D.} Perched with its bistable claw, after the flight.}
    \label{fig:fig1}
\end{figure*}

Flapping-wing flight offers significant benefits over rotor-based propulsion ranging from noise reduction, increased safety, and potentially higher maneuverability and efficiencies [1]. While these traits could grant safe flight around structures, significant challenges remain to land large flapping-wing robots on perches and branches. 

Indeed, the combination of forward flight requirement and vertical center of mass oscillations arising in large flapping-wing robots vastly increases the difficulty to perch on objects. Achieving this task - as we often see birds do seamlessly everyday - is complex in terms of the control required in stall conditions and in terms of grasping appendage capability. In a robot, this would require a vehicle capable of carrying a claw system that can hold the robot upright for manipulation, a vision system to identify the branch and a control algorithm that can reliably bring the robot there with minimal remaining velocity. We propose to solve the perching maneuver in a controlled context as a significant step towards vision-based outdoor perching. Here, we would like to show a flight experiment to demonstrate autonomous perching of flapping robots in controlled conditions. Complete information about the design, modeling and resulting flight performance of the robotic bird can be found in [2].

\section{Experiment}
Demonstration and validation of this experimental method is performed with a large 700 g, 150 cm-wingspan ornithopter whose design stems from a previous robot class described in [3]. Successful perching flights were achieved through a series of state phases, visually represented in Fig.~\ref{fig:fig1} and described in this work.
 
\subsection{Reaching flight velocity}
Large ornithopters are not capable of hovering in windless conditions, due to unfavorable scaling. Such robots can only sustain flight with forward velocity and therefore require a process to reach this velocity initially. While one can certainly achieve this velocity via a hand throw, this results in a wide range of initial conditions. We instead propose an automated launcher system, which boasts sufficient precision for repeatable flights in space-constraints, indoor conditions, see Fig.~\ref{fig:fig1}.A. 

The system features a 2 m long double rail which can accelerate a robot to 10 m/s. In our validation scenario, speeds of 4 m/s are sufficient to initiate flight without losing altitude and without excessive speed. The robotic birds are held at a perpendicular offset of 50 cm, sufficient to avoid any collisions between the wing, tail and structure. Launcher mounting of robot occurs at the trailing edge, to remain as close as possible to the center of mass. A 20 degree angle-of-attack at launch is selected. Results show that launches are consistent, both in direction and speed, allowing us to repeatedly perform controlled flights to a point location at a 14 m distance from the launcher.

\subsection{Controlled flapping flight}
At the end of the rail, the robot’s position and attitude are captured by the motion capture system and start being wirelessly streamed to the robot’s flight companion computer. The ornithopter enters a controlled flapping flight phase, see Fig.~\ref{fig:fig1}.B. This is handled by the onboard triple loop pitch-yaw-altitude controller. This flight control system needs to handle stable flight with the appendage payload, misalignment compensation system and all the onboard electronics required for flight and perch. 

The forward flight velocity is a key parameter for perching. It should be kept low to minimize impact forces and increase reaction time before impact. To achieve this, the pitch angle of the robot needs to be high. Experiments show that pitch angles above \SI{40}{\degree} result in insufficient speed and consequently loss of lateral control. As such, a pitch angle of \SI{30}{\degree}is appropriate for perching flights and is sufficient to fly stably. This set-point is maintained by a proportional-integral (PI) controller. Consequently, when the pitch reaches \SI{30}{\degree}, the velocity drops to 2.5-3 m/s, kept until perching.

The altitude target value is achieved through regulated flapping frequency control action. Using this method, all flight trajectories reached a 2 m set-point within 8-12 m of flight distance. The flight controller performs adequately for different branch altitudes, with a maximum average vertical error of 16 cm and a lateral error of rarely exceeding 60 cm. 

\subsection{Branch approach}
The claw dimension is smaller than the flight position accuracy. As this is insufficient to reliably touch the target, we propose an correction method. The leg of the robotic bird is articulated at the elbow level, permitting semi-vertical motion of the claws. When the robot gets to within 150 cm of the branch, an optical detection system is enabled, see Fig.~\ref{fig:fig1}.C. The line detection system located on the claw feeds the relative position of the branch to the leg microcontroller. The leg servo then corrects the angle of the leg to align it with the branch, at 50 Hz. During this approach phase, the flapping motion is maintained and the flight controller remains active. The perching target, a 80 cm-black-painted branch is held at 2 m height at its center by a \SI{45}{\degree} aluminum profile which reaches out through the safety net. 

\subsection{Perching}
At 20 cm from the branch, the flapping is stopped. At this point, the forward velocity of the robot is situated between 2.5 and 3 m/s. Grasping is made possible thanks to a pair of bistable claws which pivot from the open to the closed state within 25 ms and with a 2 Nm torque. A combination of soft silicone pads and spikes offers friction forces sufficient to compensate the torque resulting from center of mass offsets.  

Experiments have demonstrated that we can reach the branch location repeatedly, within the tolerance range of the leg-claw system. The low forward flight velocity shortly before landing ensures that forces remain below 150 N, significantly reducing the likelihood of damage. We report that branch perching was achieved in 6 out of 9 perching flight tests, and that no damage occurred in any of the successful flights. The results were validated with a second robot which perched without re-tuning.


\end{document}